\documentclass[a4paper,conference]{IEEEtran}
\usepackage{graphicx}

\usepackage{cite}
\ifCLASSINFOpdf
\else
\fi
\begin{document}
\title{Real-Time Drone Detection and Tracking With Visible, Thermal and Acoustic Sensors}


\author{\IEEEauthorblockN{Fredrik Svanström$^{a}$,  Cristofer Englund$^{b,c}$, Fernando Alonso-Fernandez$^{c}$}
\IEEEauthorblockA{$^{a}$ Air Defence Regiment. Swedish Armed Forces\\
$^{b}$ RISE, Lindholmspiren 3A, SE 417 56 Gothenburg, Sweden\\
$^{c}$ Center for Applied Intelligent Systems Research (CAISR), Halmstad University, SE 301 18 Halmstad, Sweden\\
Emails: DroneDetectionThesis@gmail.com, cristofer.englund@hh.se, feralo@hh.se}
}


%Telephone: (800) 555--1212, Fax: (888) 555--1212}


\maketitle

\begin{abstract}
This paper explores the process of designing an automatic multi-sensor drone detection system. % using machine learning and sensor fusion.
Besides the common video and audio sensors, the system also includes a thermal infrared camera,
which is shown to be a feasible solution to the drone detection task. Even with slightly lower resolution, the performance is just as good as a camera in visible range. The detector performance as a function of the sensor-to-target distance is also investigated.
In addition, using sensor fusion, the system is made more robust than the individual sensors, helping to reduce false detections.
%
%It is observed that when using the proposed sensor fusion approach, the results are more stable, and false detections are reduced.
%
To counteract the lack of public datasets,
a novel video dataset containing 650 annotated infrared and visible videos of drones, birds, airplanes and helicopters is also presented\footnote{https://github.com/DroneDetectionThesis/Drone-detection-dataset}.
The database is complemented with an audio dataset of the classes drones, helicopters and background noise.
\end{abstract}


% \fi
\IEEEpeerreviewmaketitle

\section{Introduction}

Small and remotely controlled unmanned aerial vehicles (UAVs), hereinafter referred to as drones, can benefit the society. Examples include delivery of defibrillators \cite{[Sanfridsson19]}, fire fighting \cite{INNOCENTE201980}, or law enforcement. Moreover, the low cost and ease of operation make drones suitable for entertainment \cite{[FAI]}.
Nevertheless, they can also be intentionally or unintentionally misused, so that safety of others can be threatened. %In the worst case,
For example, an aircraft can be severely damaged if it collides with a consumer-sized drone, even at moderate speeds %, as shown by researchers at the University of Dayton
\cite{[Dayton]}.

Due to the rapid development of commercial and recreational drones, the research area of drone detection has emerged in the last few years \cite{Taha19,[GoogleTrends]}.
%with the Internet search trend for web pages with drone detection related content increasing substantially over the last ten years \cite{[GoogleTrends]}.
%
Accordingly, this work explores the possibilities and limitations of designing and constructing an automatic multi-sensor drone detection and tracking system building on state-of-the-art machine learning techniques.
The methods will be extended from conclusions and recommendations from the related literature \cite{Guvenc18,Taha19,Guvenc18}.
Besides the necessity of effective methods for detection, classification and tracking that make use of the latest techniques, sensor fusion is indicated as an open area of importance in order to achieve more accurate results in comparison to a single sensor. However, research in this direction is scarce \cite{Samaras_2019,Guvenc18,Diamantidou19,Shi18}.
Most studies also fail to specify the type of acquisition device, the drone type, the detection range, or the employed dataset \cite{Taha19}.
The lack of proper UAV detection studies employing thermal infrared cameras is also mentioned as an issue, despite its success in detecting other type of targets \cite{Taha19}.
Also, we have not found any previous study that investigates classification performance as a function of distance to the target.
Another contribution of this work is the collection and annotation of a drone dataset to be made public, containing data from as many of the sensors as possible.
The lack of public databases of reference that could serve as bechmark for researchers is another fundamental challenge \cite{Taha19}.
%
%This will also include the collection and annotation of the necessary dataset to %accomplish the training and evaluation of the system.
%
To effectively detect the sought after drones, the system must also detect and keep track of other flying objects that are likely to be mistaken for a drone \cite{Saqib17,Aker17}.
For this purpose, we increase the number of target classes compared to previous studies.
In our research, three different consumer-grade drones are included in the dataset, together with birds, airplanes and helicopters.
%

%
%In particular, this will be achieved by: $i$) using different combination of sensors; $ii$) exploring the performance of the sensor-to-target distance; $iii$) increasing the number of target classes in comparison to previous studies; and $iv$) incorporating a novel sensor fusion method for the drone detection task.
%

\begin{figure*} [htb]
\centering
\includegraphics[width=0.82\textwidth]{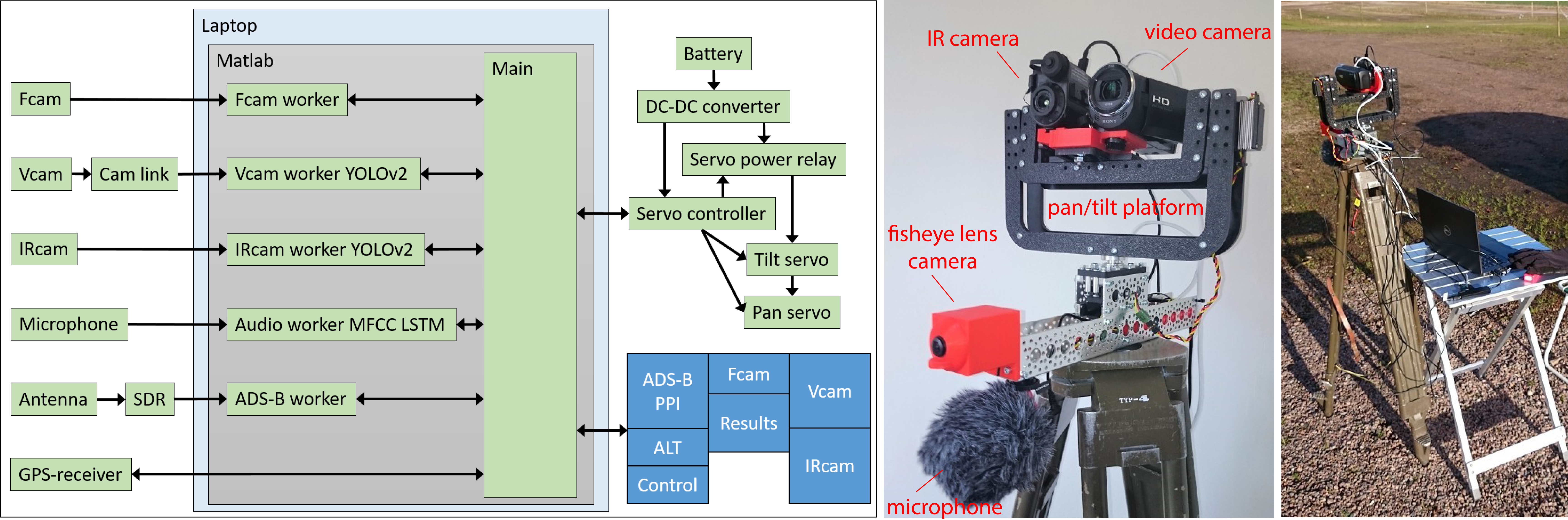}
\caption{Left: system architecture with hardware and software parts. Center: main hardware parts of the detection system. Right: the system deployed just north of the runway at Halmstad airport (IATA/ICAO code: HAD/ESMT).}
\label{fig:system}
\end{figure*}

\section{Related Works}

%An introduction to the subject of drone detection from 2018 is found in \cite{Guvenc18}.
%
%Among the identified challenges, the work mentions the development of effective methods for detection, classification and tracking. Sensor fusion is indicated as an open area of critical importance as well in order to achieve more accurate results in comparison to using a single sensor, but research in this direction is scarce \cite{Samaras_2019,Guvenc18,Diamantidou19,Shi18}.
%
%Another review publication from 2019 \cite{Taha19} stresses the lack of public datasets of reference that could serve as bechmark for researchers. Also, among the 50 references reviewed by this paper, no study investigated the classification performance as a function of drone distance. The authors also acknowledge that most of the studies fail to specify the type of acquisition device, the drone type, the detection range, or the employed dataset.
%
%Also in 2019, the review of \cite{Samaras_2019} contains 178 references, not only specific of drone detection and classification, but also regarding general aspects of machine learning or sensor fusion.
%
%Another aspect highlighted in these reviews is the lack of proper UAV detection studies employing thermal infrared cameras, despite its success in detecting other type of targets.

%\subsection{Sensors}

The sensors used for drone detection include radars, % (active and passive),
cameras in the visible spectrum, cameras detecting thermal infrared emission (IR), microphones, radio-frequency scanners to detect signals of the drone and its controller (RF), lasers (LiDAR, LADAR), %and even
humans \cite{Samaras_2019,Boddhu13},
and even animals
%. Recently it has also been successfully demonstrated that animals can be trained for this purpose
\cite{[GuardFromAboveBV]}.
%
%The fusion of multiple sensors to achieve more accurate results in comparison to %using a single sensor is also well-founded when it comes to drone detection, %since one sensor can compensate weaknesses of others \cite{Samaras_2019}.

\textbf{Thermal Infrared Sensors} are explored in  \cite{ANDRASI2017183,wang_chen_choi_kuo_2019,Diamantidou19}.
The work in \cite{ANDRASI2017183} uses a low-cost sensor providing videos of only 80$\times$60 pixels, and classification is done by a human. % looking at the stream.
The work in \cite{wang_chen_choi_kuo_2019} employs %a sensor providing
videos of 1920$\times$1080 together with deep-learning-based detection and tracking (Faster-RCNN). However, no details of the sensor are given.
To counteract the lack of thermal data, the authors %also explore the possibility of using
use a Cycle-GAN (Generative Adversarial Network) \cite{[Zhu17cycleGAN]} to produce synthetic training images. % thermal training data.
A thermal camera is also used in \cite{Diamantidou19}, without further details regarding the type, field of view (FoV), or resolution.

\textbf{Sensors in the Visible Range} are the most widespread used, combined with deep-learning methods.
The work \cite{Park17} compares six different CNN models, concluding that YOLO v2 \cite{Redmon17} might be the most appropriate, considering speed and accuracy trade-offs.
YOLO v2 is the preferred choice in many works \cite{Wu18,Liu_2018,Saqib17,Aker17}.
A lightweight version, the more recent YOLO v3, is used in \cite{Unlu19}.
The use of pan/tilt platforms to steer cameras in the direction of targets has also lead to the use of wide-angle sensors.
In \cite{Unlu19}, a camera with 110$^{\circ}$ FoV is used to align a rotating narrow-field camera. To find the objects to be investigated by the narrow-field camera, the stream of the wide-angle camera is analysed by a Gaussian Mixture Model (GMM) foreground detector.
Among the mentioned papers, \cite{Liu_2018} has the biggest amount of target classes (drone, airplane, helicopter), followed by \cite{Aker17} with two (drone, bird). %None of the papers report the detection accuracy as a function of the sensor-to-target distance.

\textbf{Acoustic Sensors} are also explored by numerous papers \cite{Kim17,Siriphun18,Park15,Anwar19,Liu17,Jeon17,Bernardini17,Busset15}.
Some \cite{Kim17,Siriphun18,Park15} use the Fast Fourier Transform (FFT) for feature extraction.
However, Mel Frequency Cepstrum Coefficients (MFCC) are the most common features \cite{Anwar19,Liu17,Jeon17,Bernardini17}.
The paper \cite{Jeon17} compares three methods, concluding that Long Short-Term Memory (LSTM) networks get the best performance.
In \cite{Jeon17}, the classification is binary (drone/ background), that the present paper extends with a helicopter class.
The maximum detection range reported in these papers is 290m \cite{Busset15} with a 120-element microphone array.

\textbf{Radar} is the most common technology to detect flying objects. However, detecting drones with systems designed for aircrafts is not straightforward because they often use techniques to reduce unwanted echoes from small, slow and low-flying objects, which is precisely what characterise drones.
Also, the Radar Cross Sections (RCS) of medium-sized consumer drones are similar to birds, which can lead to false targets \cite{Gong19,Patel18,Herschfelt17}.
Nevertheless, several works have explored the use of micro-doppler characteristics of drones \cite{Fuhrmann17,Bjorklund18,Drozdowicz16,Rahman18}.
Typically, echoes from the %moving blades of the
propellers are used for detection, on top of the bulk motion doppler signal of the drone.

\textbf{Other detection Techniques} include the RF fingerprint of the drone or its controller \cite{Birnbach17,Shorten18,Ezuma20}, and lasers like LiDAR and LADAR %(Light/LAser Detection And Ranging)
\cite{Kim_2018}. In \cite{Shorten18}, a CNN is used with an antenna array to calculate the direction to the controller with a precision of a few degrees. In \cite{Ezuma20}, signals from 15 different controllers are classified with an accuracy of 98.13\% with only three RF-features and a K-Nearest Neighbour classifier.
LiDAR and LADAR are used in \cite{Kim_2018} in combination with background substraction to detect drones on distances up to 2 km.

\begin{figure*} [htb]
\centering
\includegraphics[width=0.85\textwidth]{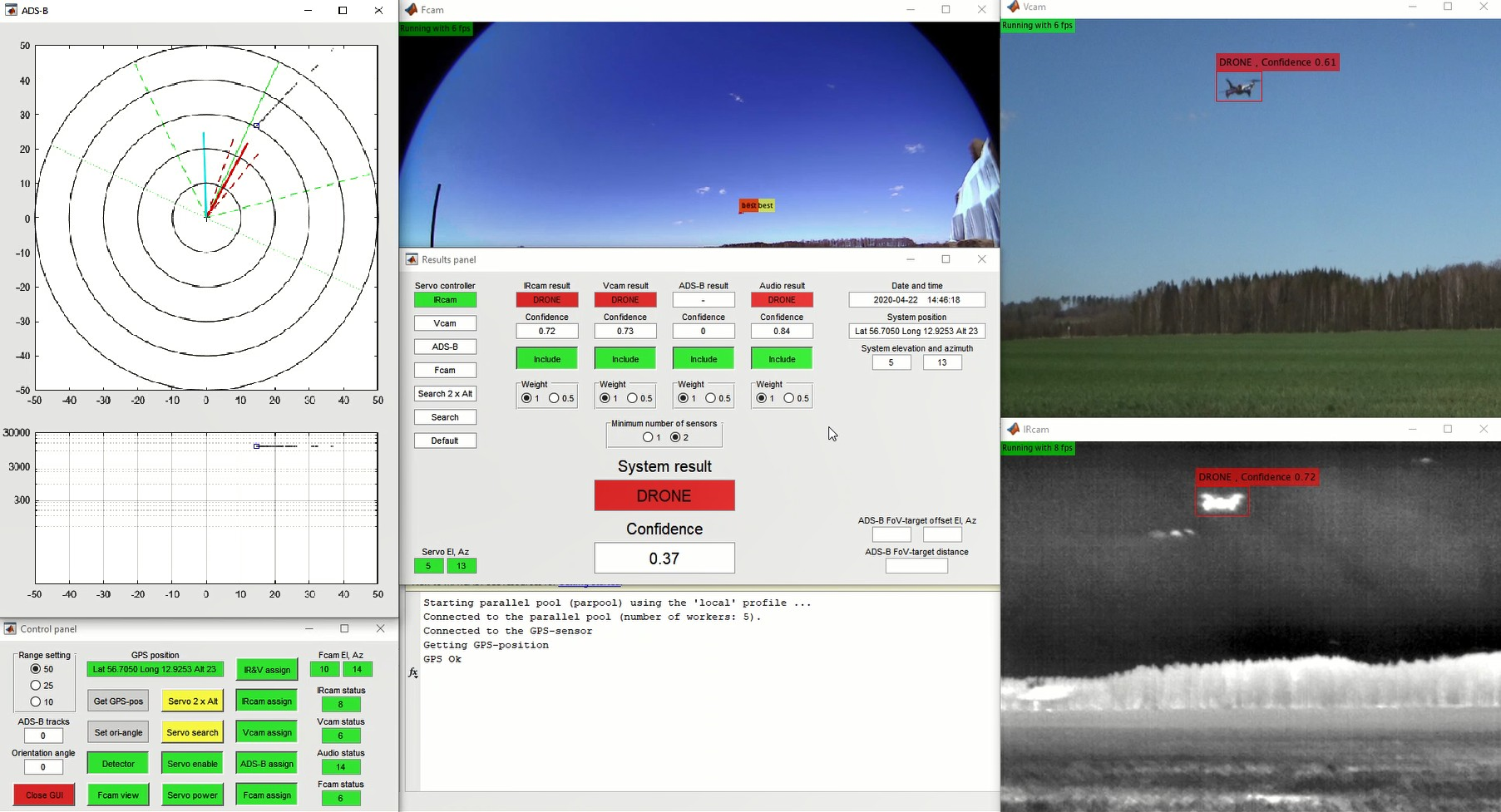}
\caption{Graphical User Interface. See Section~\ref{sect:GUI} for further details.}
\label{fig:system_GUI}
\end{figure*}

\section{Methods and Materials}

%%FOR JOURNAL
%This work can be seen as incorporating all phases of %designing a drone detection and classification system. From %the initial literature study, the subsequent assessment and %procurement of suitable sensors and hardware, the design and %3D-printing of parts that are not available, the programming %and training of the system, and finally the evaluation.
%

This section describes the automatic drone detection system, first on a system-level and thereafter in detail, both regarding hardware and software.
%
%We also include the collection of the required dataset.
%
This emerges from previous studies \cite{Taha19} which indicate that the majority of works fail to specify the acquisition device, drone type, detection range, or dataset.

\subsection{System Architecture}

An efficient detection system must have the capability to both cover a large volume of airspace, and have enough resolution to distinguish the drone from other objects.
Two ways of accomplishing this is combining wide and narrow FoV cameras \cite{Unlu19}, or using an array of high-resolution cameras \cite{Liu_2018}.
We follow the first approach, since our solution is intended to be portable, and coverage will be achieved by a moving pan/tilt platform.
To have the ability to track moving objects, the system must work in real time, with the feedback loop of the moving platform at sub-second speed.
This in turn imposes demands in the efficiency of software.
The system architecture, including the principal layout (in blue) of the Graphical User Interface (GUI) is shown in Figure~\ref{fig:system}, left.

\subsection{Hardware}

The system uses a thermal infrared camera (IRcam) and a video camera in visible range (Vcam) as primary sensors for detection.
To keep track of cooperative aircrafts, ADS-B information is made available with an antenna. Cooperative aircraft are defined as equipped with a system that broadcasts the aircraft’s position, velocity vectors and identification information.
Audio data is used to detect if a drone or a helicopter is in the vicinity using their respective distinctive sounds.
Due to the limited FoV of the primary cameras, they are steered towards specific directions guided by a fish-eye lens camera (Fcam) covering 180$^{\circ}$ horizontally and 90$^{\circ}$ vertically.
The role of the fish-eye camera is not to detect specific classes, but to detect moving objects in its field of view.
If nothing is detected by the Fcam, the platform can be set to move in two different search patterns to scan the sky around the system.
All computations are made in a standard laptop, which is also used to present the results.
The main hardware parts are shown in the center of Figure~\ref{fig:system}. %The holder for the servo controller and power relay boards is placed %behind the pan servo inside the aluminium mounting channel.
To have a stable base, all hardware components, except the laptop, are mounted on a standard surveyors tripod. This also facilitates transport and deployment of the system outdoors, as shown in the right part of the figure.

The \textbf{thermal infrared} camera is a FLIR Breach PTQ-136 using the Boson 320$\times$256 pixels detector (Y16 with 16-bit grey scale). The FoV is 24$^{\circ}$ horizontally and 19$^{\circ}$ vertically.
Figure~\ref{fig:system_GUI}, bottom right, shows an example image.
Notably, this sensor has a higher resolution than the one used in \cite{ANDRASI2017183} where a FLIR Lepton sensor with 80$\times$60 pixels was used.
Also, in \cite{ANDRASI2017183} the authors were able to detect three drone types up to a distance of 100m, but done manually by a person looking at the live video stream, in contrast to the present paper where we will use an automatic system.
The output is sent to the laptop via a USB-C port at 60 frames per second (FPS). The IRcam is also powered by the USB connection.
A Sony HDR-CX405 video camera is used to record in the \textbf{visible range}.
The output is a HDMI-signal, hence an Elgato Cam Link 4K frame grabber is used to feed the laptop with a 1280$\times$720 video stream in YUY2-format (16 bits per pixel) at 50 FPS.
The Vcam has an adjustable zoom lens so the FoV can be both wider or narrower than the IRcam. However, it is set to have about the same FoV as the IRcam.
To  monitor a larger part of the surroundings, an ELP 8 megapixel 180$^{\circ}$ \textbf{fish-eye lens} camera is also used. It outputs a 1024$\times$768 video stream in Mjpg-format at 30 FPS via USB.
To be able to capture the distinct sound that drones emit when flying, a Boya BY-MM1 mini cardioid directional \textbf{microphone} is also connected to the laptop.
To track aircraft equipped with transponders, an \textbf{ADS-B receiver} is also used. It consists of an antenna and a NooElec Nano 2+ Software Defined Radio receiver (SDR).
%This is tuned to 1090 MHz so that the identification and positional data sent out as a part of %the 1 Hz squitter message can be decoded and displayed.
The Nano 2+ SDR receiver is connected by USB.
To present the decoded ADS-B data in a correct way the system is also equipped with a G-STAR IV BU-353S4 \textbf{GPS receiver} connected via USB. The receiver outputs messages following the National Marine Electronics Association (NMEA) format standard.
To be able to detect targets in a wider field of view than just the 24$^{\circ}$ horizontally and 19$^{\circ}$ vertically of the IRcam and Vcam, these are mounted on a \textbf{pan/tilt platform}.
This is the Servocity DDT-560H direct drive tilt platform together with the DDP-125 Pan assembly, also from Servocity. To achieve the pan/tilt motion two Hitec HS-7955TG servos are used.
A Pololu Mini Maestro 12-Channel USB servo controller is included so that the respective position of the servos can be controlled from the laptop.
%
%Since the servos have shown a tendency to vibrate when holding the platform in specific directions, a third channel of the servo controller is also used to give the possibility to switch on and off the power to the servos using a small optoisolated relay board.
%
To supply the servos with the necessary voltage and power, both a net adapter and a DC-DC converter are available. The DC-DC solution is used when the system is deployed outdoors and, for simplicity, it uses the same battery type as one of the available drones.
The computational part is handled by a Dell Latitude 5401 laptop equipped with an Intel i7-9850H CPU and an Nvidia MX150 GPU. All the mentioned sensors and servo controller are connected using the built-in USB ports and an additional USB-hub, as shown in Figure~\ref{fig:system}, right.

\subsection{Software}

The software used can be divided into two parts. First, the software running in the system when it is deployed. Additionally, there is a support software to form the training data sets and to train the system.

The software running when the drone detector is deployed consists of the main script and five `workers', as shown in Figure~\ref{fig:system}, left.
The main script and the workers are set up so that they can be run independently of each other in a stand-alone mode.
These are threads running in parallel, enabled by the Matlab parallel computing toolbox.
This also allows the different detectors to run asynchronously, handling as many frames per second as possible without any inter-sensor delays and waiting time.
Transfer of messages between the main program and the workers is done using pollable data queues.
The \textbf{main script} communicates with the servo controller and the GPS receiver. At a regular frequency of 10Hz, it also interacts with the workers and the servo controller to read and update the platform position, and to update the GUI.
The \textbf{Fcam worker} utilizes a foreground/background detector
based on Gaussian Mixture Models (GMM), which produces binary masks of moving objects.
This is followed by a multi-object Kalman filter tracker which, after calculating the position of the best-tracked target, sends the azimuth and elevation angles to the main program.
The best-tracked target is defined as the one with the longest track history.
Based on this, the main program can then control the pan/tilt platform servos via the servo controller, so that the moving object can be analysed further by the infared and visible cameras.
The \textbf{IRcam and Vcam workers} are similar in their basic structure, running a trained YOLO v2 detector and classifier. The information sent to the main script is the class of the detected target, the detection confidence, and the horizontal and vertical offsets in degrees from the centre of the image. The latter information is used by the main script to calculate servo commands when an object is being tracked by the system.
YOLO provides an array of class labels, detection confidence, and bounding boxes of detected objects.
Since it predicts multiple bounding boxes for the same object, only the strongest one is given, which is chosen as the box with the highest IoU (Intersection Over Union) with the annotations in the training data.
To be assigned to a class, the chosen bounding box should also have a minimum IoU with the training object that it is supposed to detect.
Looking at the related works, an IoU of 0.5 is usually employed \cite{Park17,Saqib17,Aker17}.
A threshold to the detection confidence can be also imposed, so bounding boxes with small confidence can be rejected, even if their IoU is above 0.5.
The \textbf{Audio worker} collects acoustic data in a one-second long buffer at 44100 Hz, which is updated 20 times per second. To classify the source in the buffer, it is first processed by extracting MFCC features, which are then sent to a LSTM classifier.
The worker then sends information about the class and confidence to the main script.
Unlike the others, the \textbf{ADS-B worker} has two output queues. One consists of current tracks and the other of the history tracks. This is done so that the presentation clearly shows the heading and altitude changes of the targets.
All of the above workers also send a confirmation of the command from the main script to run the detector/classifier or to be idle. The number of frames per second currently processed is also sent to the main script.
Table~\ref{tab:output-classes}
shows the different output classes that the main program can receive from the workers.
Note that not all sensors can output all the target classes. The audio worker has an additional background class, and the ADS-B will output the `no data' class if the vehicle category field of the received message is empty.

\begin{table}[b]
%\small
%\footnotesize
%\scriptsize
\begin{center}
\begin{tabular}{c|ccccc}

\multicolumn{6}{c}{} \\

sensor & \multicolumn{5}{c}{output classes}   \\ \hline \hline

\textbf{IRcam} & airplane & bird & drone  & helicopter  &   \\ \hline
\textbf{Vcam} & airplane & bird & drone  & helicopter  &   \\ \hline
\textbf{Audio} &  &  & drone  & helicopter  &  background \\ \hline
\textbf{ADS-B} & airplane & & drone & helicopter  &  no data  \\ \hline

\end{tabular}

\end{center}
\caption{Output classes of the sensors.}
\label{tab:output-classes}
\end{table}
\normalsize

\begin{table}[b]
%\small
%\footnotesize
%\scriptsize
\begin{center}
\begin{tabular}{ccccc}

%\multicolumn{5}{c}{} \\

\multicolumn{1}{c}{} & \multicolumn{4}{c}{IR videos (365)}   \\  \cline{2-5}

bin & airplane & bird & drone  & helicopter    \\ \hline
Close & 9 & 10 & 24  & 15   \\
Medium & 25 & 23 & 94 & 20 \\
Distant & 40 & 46 & 39 & 20 \\ \hline

\multicolumn{5}{c}{} \\

\multicolumn{1}{c}{} & \multicolumn{4}{c}{visible videos (285)} \\  \cline{2-5}

bin & airplane & bird & drone  & helicopter    \\ \hline
Close & 17 & 10 & 21  & 27   \\
Medium & 17 & 21 & 68 & 24  \\
Distant & 25 & 20 & 25 & 10   \\ \hline

\end{tabular}

\end{center}
\caption{Distribution of the IR and visible videos.}
\label{tab:db-stats}
\end{table}
\normalsize

\begin{figure} [t]
\centering
\includegraphics[width=0.4\textwidth]{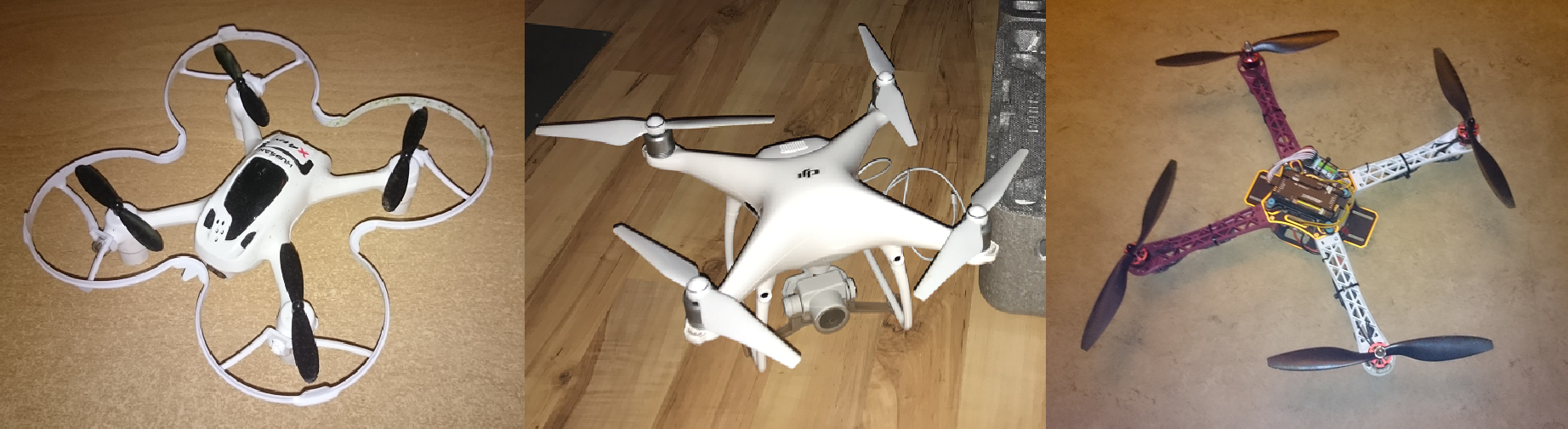}
\caption{The three drones of our dataset. Left: Hubsan H107D+. Middle: DJI Phantom 4 Pro. Right: DJI Flame Wheel F450.}
\label{fig:db_drones}
\end{figure}

\begin{figure} [t]
\centering
\includegraphics[width=0.4\textwidth]{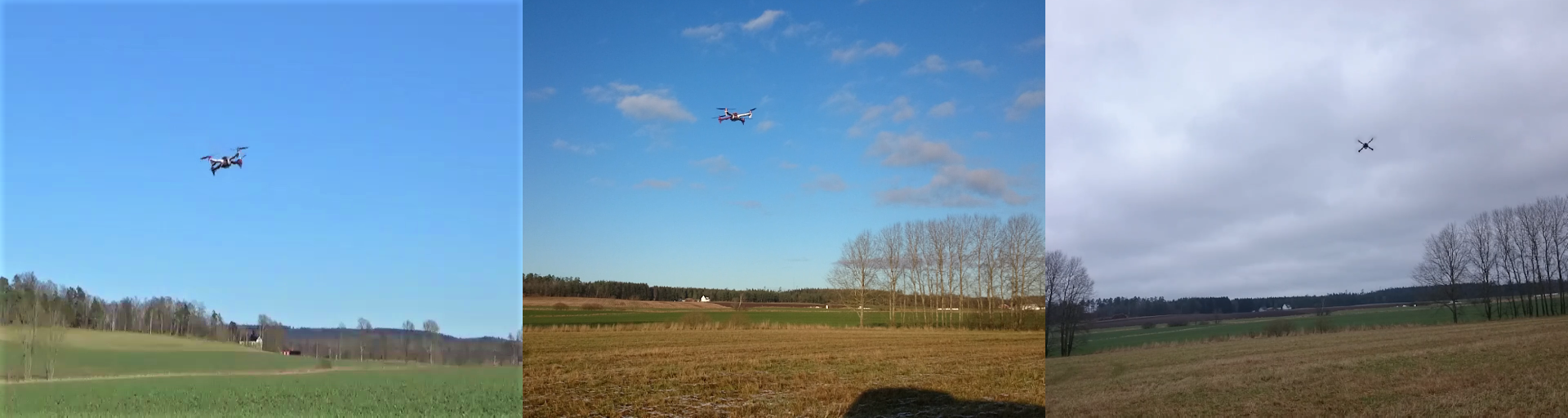}
\caption{Examples of varying weather conditions in the dataset.}
\label{fig:db_weather}
\end{figure}

\subsection{Graphical User Interface}
\label{sect:GUI}

Figure~\ref{fig:system_GUI} shows the GUI of the system. It presents the results of the different sensors/workers, and also provides possibility of easily controlling the system.
The top-left part is the ADS-B presentation area, consisting of a PPI-type display (Plan Position Indicator) and an altitude display.
The green lines present the field of motion of the pan/tilt platform (dashed) and the field of view of the fish-eye lens camera (dotted).
The actual direction of the pan/tilt platform is presented with a solid red line and the field of view of the IR- and video cameras are represented using dashed red lines.
If any object is tracked by the fish-eye lens camera worker its direction is indicated by a solid cyan line, which also shows the track history plots.
The presentation of the altitude information is done in a logarithmic plot so that the lower altitude portion is more prominent.
The area below the ADS-B presentation area is the control panel, containing buttons to control different settings of the ADS-B receiver, GPS receiver, and servos of the pan/tilt platform.
It also provides the status information of the different sensors with red/yellow/green colours.
Streams captured by the Fcam (fish-eye), Vcam (visible) and IRcam (thermal infrared) are shown respectively in the top center, top right, and bottom right.
In the bottom center, the Matlab command window (primary development environment) is shown, so messages like exceptions or errors can be monitored during the development and operation of the system.
Finally, the results panel appears in the center, presenting results of the workers and other information such as the data source (camera) currently controlling the servos of the pan/tilt platform, the angles of the servos, current time, GPS position, elevation and azimuth relative to the north, and ADS-B targets detected.

\section{Drone Detection Dataset}

A dataset has been captured at three airports in Sweden: Halmstad Airport (IATA/ICAO code: HAD/ESMT), Gothenburg City Airport (GSE/ESGP) and Malmö Airport (MMX/ESMS).
Three different drones are used (Figure~\ref{fig:db_drones}): Hubsan H107D+, a small-sized first- person-view (FPV) drone; the high-performance DJI Phantom 4 Pro; and the medium-sized DJI Flame Wheel. The latter can be built both as a quadcopter (F450) or in a hexacopter configuration (F550). The version used is an F450 quadcopter.
They differ a bit in size, with Hubsan H107D+ being the smallest (side length from motor-to-motor of 0.1 m). The Phantom 4 Pro and the DJI Flame Wheel F450 have 0.3 and 0.4 m motor-to-motor side length, respectively.
Flights are done in compliance with the national rules for unmanned aircrafts.
Since the drones must be flown within visual range, the dataset is recorded in daylight, even if the system can be used at night using the thermal infrared and acoustic sensors. The ADS-B information received is naturally also working at night.
The weather in the dataset stretches from clear and sunny, to scattered clouds and completely overcast, as shown in Figure~\ref{fig:db_weather}.

Both the videos and the audio-files are cut into ten-second clips to be easier to annotate. To obtain a more comprehensive dataset, both in terms of aircraft types and sensor-to-target distances, our data has been complemented with non-copyrighted material from the YouTube channel `Virtual Airfield operated by SK678387' \cite{[VIRTUALAIRFIELD]}, in particular 11 plus 38 video clips in the airplane and helicopter categories, respectively.
This is because it has not been possible to film all types of suitable targets, given that this work has been carried our during the drastic reduction in flight operations due to the COVID19 pandemic.
Overall, the dataset contains 90 audio clips and 650 videos (365 IR and 285 visible, of ten seconds each), with a total of 203328 annotated images, and it has been made publicly available.
The IR videos have a resolution of 320$\times$256 pixels, whereas the visible videos have 640$\times$512.
The greatest sensor-to-target distance for a drone in the dataset is 200 m.
The audio part has 30 ten-second clips of each of the three output audio classes (Table~\ref{tab:output-classes}),
whereas the distribution of videos among the four output video classes is shown in Table~\ref{tab:db-stats}.
The background sound class contains general background sounds recorded outdoor in the typical deployment environment of the system, and also includes some clips of the sounds from the servos moving the pan/tilt platform.

Since one of our objectives is to explore performance as a function of the sensor-to-target distance, the video set has been divided into three category bins: Close, Medium and Distant. The borders between them are chosen to follow the industry-standard Detect, Recognize and Identify (DRI) requirements \cite{[DRI]}, building on the Johnson criteria \cite{[Chevalier16]}.
The Close bin is from 0 m out to a distance where the target is 15 pixels wide in the IRcam image (requirement for `identification' of the target according to DRI, e.g. the specific drone model). The Medium bin stretches from where the target is from 15, and down to 5 pixels (requirement for `recognition' of the target, e.g. a drone and not another object, albeit without the possibility of identifying the model), and the Distant bin is beyond that (requirement for `detection', e.g. there is something).
%
%Given the resolution and field of view of the IRcam and the object class %sizes (drone 0.4 m, bird 0.8 m, helicopter 10 m, airplane2 20 m), we get a %distance division for the different object types summarized in Table 3.
%
%To illustrate the detect, recognize and identify concept, objects from all %the target classes of our database being 15 pixels in width are shown in %Figure XX.

\begin{table}[b]
%\small
%\footnotesize
%\scriptsize
\begin{center}
\begin{tabular}{cccccc}

%\multicolumn{6}{c}{} \\

\multicolumn{1}{c}{} & \multicolumn{5}{c}{distance bin: CLOSE}   \\ \cline{2-6}

 & airplane & bird & drone  & helicopter  &  average \\ \hline

\textbf{Precision} & 0.9197 & 0.7591 & 0.9159 & 0.9993& 0.8985 \\
\textbf{Recall} & 0.87367 & 0.85087 & 0.87907 & 0.87927 & 0.8706   \\ \hline
\textbf{F1-score} & & & & & 0.88447   \\
\hline

\multicolumn{6}{c}{} \\

\multicolumn{1}{c}{} & \multicolumn{5}{c}{distance bin: MEDIUM}   \\ \cline{2-6}

 & airplane & bird & drone  & helicopter  &  average \\ \hline

\textbf{Precision} &  0.82817 & 0.50637 & 0.89517 & 0.95547 & 0.7962  \\
\textbf{Recall} &  0.70397 & 0.70337 & 0.80347 & 0.83557 & 0.7615 \\ \hline
\textbf{F1-score} & & & & & 0.77857   \\ \hline

\multicolumn{6}{c}{} \\

\multicolumn{1}{c}{} & \multicolumn{5}{c}{distance bin: DISTANT}   \\ \cline{2-6}

 & airplane & bird & drone  & helicopter  &  average \\ \hline

\textbf{Precision} &  0.78227 & 0.61617 & 0.82787 & 0.79827 & 0.7561  \\
\textbf{Recall} &  0.40437 & 0.74317 & 0.48367 & 0.45647 & 0.5218 \\ \hline
\textbf{F1-score} & & & & & 0.61757   \\ \hline

%\multicolumn{6}{c}{} \\

%%\multicolumn{1}{c}{} & \multicolumn{4}{c}{distance bin} & %%\multicolumn{1}{c}{} \\ \cline{2-5}

% & CLOSE & MEDIUM & DISTANT  &  average  &   \\ \cline{1-5}

%\textbf{F1-score} & 0.88447 & 0.77857 & 0.61757 & 0.7601 \\ \cline{1-5}

\end{tabular}

\end{center}

%\begin{center}
%\begin{tabular}{ccccc}
%
%\multicolumn{5}{c}{} \\
%
%\multicolumn{1}{c}{} & \multicolumn{4}{c}{distance bin}   \\ \cline{2-5}
%
% & close & medium & distant  &  average \\ \hline
%\textbf{F1-score} &  &  &   &   &   \\ \hline
%
%
%\end{tabular}
%
%\end{center}

\caption{Results with the thermal infrared sensor. The average of the three F1-scores is 0.7601}
\label{tab:results-IRcam}
\end{table}
\normalsize

\begin{table}[b]
%\small
%\footnotesize
%\scriptsize
\begin{center}
\begin{tabular}{cccccc}

%\multicolumn{6}{c}{} \\

\multicolumn{1}{c}{} & \multicolumn{5}{c}{distance bin: CLOSE}   \\ \cline{2-6}

 & airplane & bird & drone  & helicopter  &  average \\ \hline

\textbf{Precision} & 0.8989 & 0.8284 & 0.8283 & 0.9225 & 0.8695 \\
\textbf{Recall} & 0.7355 & 0.7949 & 0.9536 & 0.9832 & 0.8668 \\ \hline
\textbf{F1-score} & & & & & 0.8682   \\
\hline

\multicolumn{6}{c}{} \\

\multicolumn{1}{c}{} & \multicolumn{5}{c}{distance bin: MEDIUM}   \\ \cline{2-6}

 & airplane & bird & drone  & helicopter  &  average \\ \hline

\textbf{Precision} & 0.8391 & 0.7186 & 0.7710 & 0.9680 & 0.8242 \\
\textbf{Recall} & 0.7306 & 0.7830 & 0.7987 & 0.7526 & 0.7662 \\ \hline
\textbf{F1-score} & & & & & 0.7942   \\ \hline

\multicolumn{6}{c}{} \\

\multicolumn{1}{c}{} & \multicolumn{5}{c}{distance bin: DISTANT}   \\ \cline{2-6}

 & airplane & bird & drone  & helicopter  &  average \\ \hline

\textbf{Precision} & 0.7726 & 0.6479 & 0.8378 & 0.6631 & 0.7303 \\
\textbf{Recall} & 0.7785 & 0.7841 & 0.5519 & 0.5171 & 0.6579 \\ \hline
\textbf{F1-score} & & & & & 0.6922   \\ \hline

\end{tabular}

\end{center}

\caption{Results with the visible camera. The average of the three F1-scores is 0.7849}
\label{tab:results-Vcam}
\end{table}
\normalsize

\begin{table}[t]
%\small
%\footnotesize
%\scriptsize
\begin{center}
\begin{tabular}{ccccc}

%\multicolumn{6}{c}{} \\

\multicolumn{5}{c}{}    \\ \cline{2-5}

 & drone & helicopter & background  & average \\ \hline

\textbf{Precision} & 0.9694 & 0.8482 & 0.9885 & 0.9354 \\
\textbf{Recall} & 0.9596 & 0.9596 & 0.8687 & 0.9293 \\ \hline
\textbf{F1-score} & & & & 0.9323   \\
\hline

\end{tabular}

\end{center}

\caption{Results with the audio detector.}
\label{tab:results-audio}
\end{table}
\normalsize

\section{Results}

First, the evaluation is done measuring performance in terms of precision, recall and F1-score of the individual sensors.
Secondly, the evaluation is done after sensor fusion, a scarce direction in the literature \cite{Diamantidou19,Shi18}.
From the video dataset, 120 IR and 120 visible clips (5 from each class and target bin per spectrum) are put aside to form the evaluation dataset. Out of the remaining videos, 240 are then picked as evenly distributed as possible to create the training set.
The evaluation set for the audio classifier contains five 10-second clips from each output category, and the remaining clips for training.

\subsection{Thermal Infrared Sensor (IRcam)}

As mentioned, the IRcam worker uses YOLO v2 as detector. The input layer and number of output classes are updated to the values of our database (256$\times$256 input images, 4 classes).
The detector is trained with the 120 available clips of about 10 seconds each (37428 frames in total) during 5 epochs in a computer with an Nvidia GeForce RTX2070 8GB GPU. SGDM is used as optimizer with an initial learning rate of 0.001.
Results of the IRcam detector are shown in Table~\ref{tab:results-IRcam} for each distance bin
(confidence detection threshold and IoU requirement of 0.5 in both).
Taking the average results of each distance bin, the corresponding F1-scores are also shown.
It can observed that precision and recall are well balanced within each distance bin.
As it can be expected, results are worse as the sensor-to-target distance increases.
Altering the detection threshold to a higher value (e.g. 0.8) produces an increase in the precision, at the cost of a lower recall, which is more prominent as the sensor-to-target distance increases (results not shown).
Also, from observations, a common source of false alarms of the IRcam detector is small clouds and edges of large clouds lit up by the sun.
%An example of this can be seen in Figure XX.

\subsection{Visible Camera (Vcam)}

To be able to compare the results of the IRcam and Vcam workers, the same methods and settings as above are used.
Given the bigger size of VCam images, the only difference is the input image size employed with YOLO (416$\times$416).
The training set here consists of 37519 images.
Results of the Vcam detector are shown in Table~\ref{tab:results-Vcam}
(confidence detection threshold and IoU requirement of 0.5).
These results differ no more than 3\% from the IRcam detector. Recall that the input image size of the Vcam worker is 1.625 higher, so even with a lower resolution and the fact that images of the IRcam are in greyscale, the IRcam sensor performs as well as the visible one.
This conforms well with the conclusions in \cite{ANDRASI2017183}, where the IR detector outperforms the visible one when the image size is the same.
In \cite{Park17},
%Comparing the results to other papers we see that
the YOLO v2 detector
%in \cite{Park17}
achieves an F1-score of 0.728 with exactly the same detection threshold and IoU-requirement. This is just below our results with the IRcam and Vcam workers.
However, one notable difference lies in that the detector in \cite{Park17} has only one output class. There is no mention in \cite{Park17} either to the input image size or to the sensor-to-target distance, other than 75\% of the drones have a width smaller than 100 pixels.
Also, from observations of our results, the most common source of problems is the autofocus feature of the video camera. In this setting, clear skies are not ideal, but rather a scenery with objects that can help the camera to set the focus properly.

%\subsection{Fish-Eye Lens Camera (Fcam)}

%Not sure how to include the Fcam in the results since the videos in the database (IR, visible) I presume that are with zoomed images. Do we have fish-eye images to do a quantitative evaluation?

%Also, it is said that the Fcam do not work beyond 50 m for drones (which is the beginning of the distant bin) so I am not sure if mentioned that would be counterproductive

\begin{figure} [htb]
\centering
\includegraphics[width=0.4\textwidth]{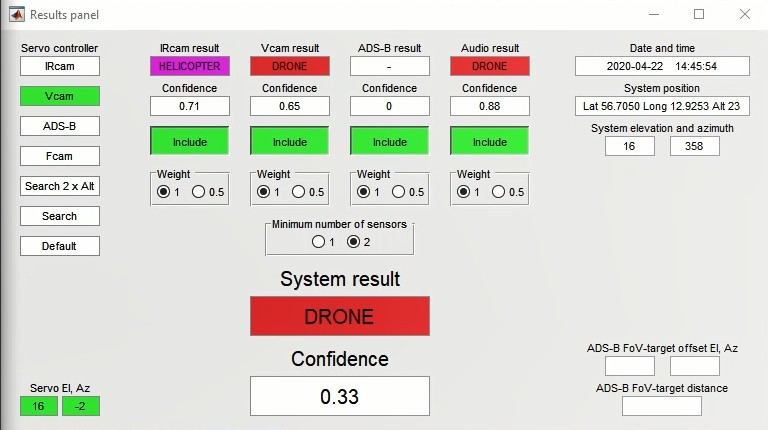}
\includegraphics[width=0.4\textwidth]{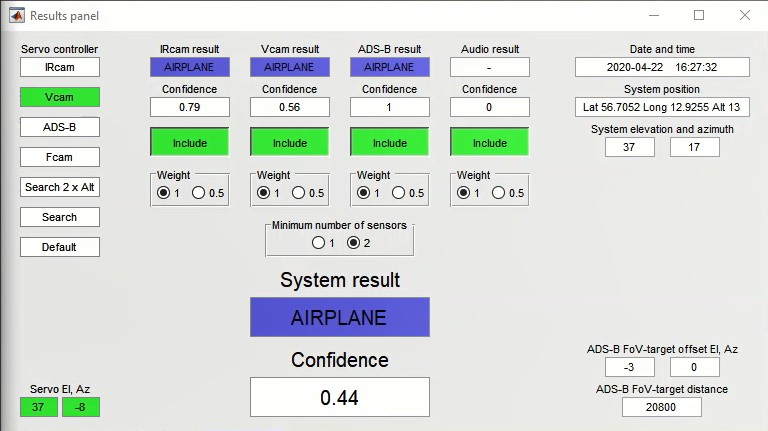}
\caption{Two examples of sensor fusion results.}
\label{fig:fusion-options}
\end{figure}

\subsection{Audio Sensor}

The audio worker uses MFCC features and a LSTM classifier to identify the source of the sound.
The LSTM classifier builds on \cite{[LSTM]}, but increasing the output classes from two to three.
It is trained from scratch during 120 epochs with SGDM as optimizer and a learning rate of 0.001.
Results of the acoustic detector are shown in Table~\ref{tab:results-audio}.
Our F1-score is higher compared to \cite{Jeon17}. The classifier in that paper also utilize MFCC features, and out of the three network model types tested, the one comprising a LSTM-RNN performs the best, with a F1-score of 0.6984. The classification problem in \cite{Jeon17} is binary (drone or background).
Another paper using MFCC features is \cite{Bernardini17}. Using a Support Vector Machine (SVM) classifier, the authors report a precision of 0.983. Five output classes are used (drone, nature daytime, crowd, train passing and street with traffic), and the classification is based on a one-against-one strategy.
%
%From observations, the practical range of our acoustic system against a %drone is 35-45 m, depending on how the drone is flying. This is in parity %with the 50 m of [23], but far from the 160 m against a F450 drone %reported in \cite{Bernardini17} with a much more complex microphone configuration (a 120 %elements microphone array).
%
%The classification range of the system against helicopters has not been tested practically because such sounds in our database are from YouTube videos, and not recorded by us.

\begin{figure} [htb]
\centering
\includegraphics[width=0.12\textwidth]{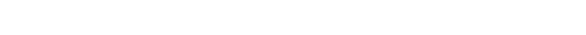}
\includegraphics[width=0.12\textwidth]{blank.png}
\includegraphics[width=0.12\textwidth]{blank.png}
\includegraphics[width=0.35\textwidth]{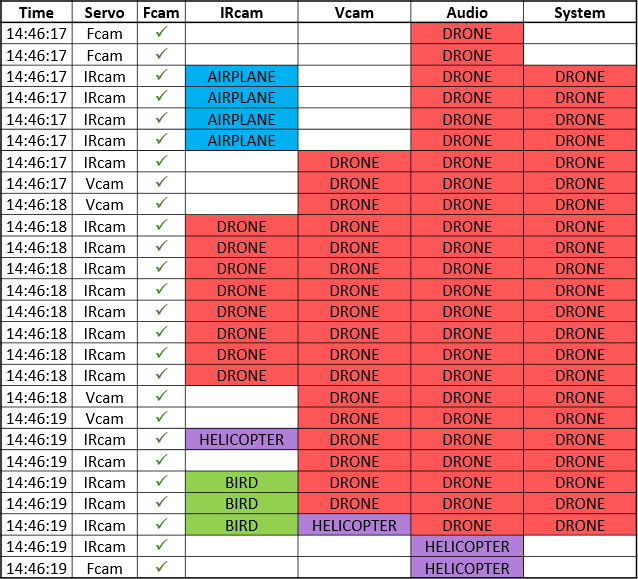}
\caption{Frame-by-frame analysis of drone detection during one evaluation session.}
\label{fig:fusion-results1}
\end{figure}

\begin{figure} [htb]
\centering
\includegraphics[width=0.35\textwidth]{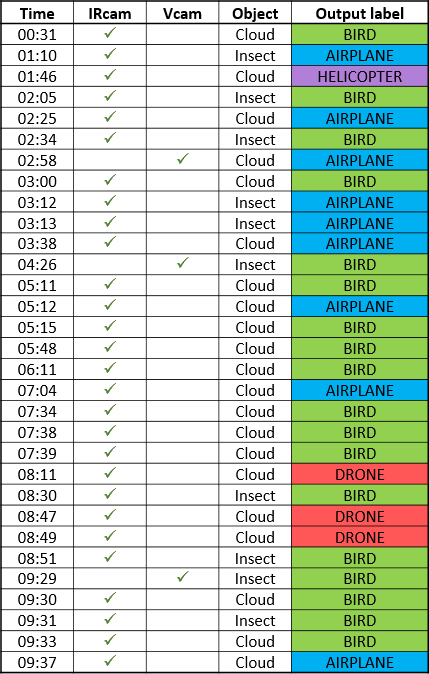}
\caption{False detections appearing in a ten minutes long section of screen recording from an evaluation session, including the type of object causing the false detection.}
\label{fig:fusion-results2}
\end{figure}

\subsection{Sensor Fusion}

The sensor fusion method employed is to utilize the class outputs and the confidence scores of the available sensors, and also to smooth the result over time (about one second).
With the dynamical setting available in the GUI, it is possible to use not only the OR-function, as in \cite{Shi18}, but more sophisticated variants by varying the number of sensors included and required for detection, including the weights for them.
Figure~\ref{fig:fusion-options}
shows an example of how sensor fusion is enabled in the GUI, which includes choice of the weight for each sensor, and the minimum number of sensors required.
Evaluating the fusion has been harder than expected due to the cancellation of flights in the airports where we captured our database during the COVID19 pandemic.
This decreased drastically the possibility for a thorough system evaluation against airplanes.

Using the screen recorder, it is possible to do a frame-by-frame analysis of a typical drone detection. An example of this is found in Figure~\ref{fig:fusion-results1}.
The servo column indicates the current servo controlling sensor. The next column specifies if the Fcam motion detector is tracking the drone or not. The respective output labels are shown in the rest of the columns. Note that the system output is more stable and lasts for more frames than the IRcam and Vcam individually, indicating the benefit of the sensor fusion.
Figure~\ref{fig:system_GUI} is the third frame from 14:46:18. Here the IRcam, Vcam and audio workers all detect and classify the drone correctly. The Fcam worker is also tracking the drone, and the pan/tilt platform at this moment is being controlled by the IRcam worker.
Comparing the system results after the sensor fusion (sensor weight 1.0   and minimum number of sensors set to two) with the output from the respective sensors, we can observe that the system outputs a drone classification at some time in 78\% of the detection opportunities. Closest to this is the performance of the Vcam detector that outputs a drone classification in 67\% of the opportunities.

It is also possible to look at the system behaviour without a drone flying in front of it. This provides an opportunity to analyse the false detections that the system outputs. Out of the videos from the evaluation session, a ten minutes long section was reviewed frame-by-frame. Figure~\ref{fig:fusion-results2} shows the timestamps, sensor types, causes of the false detection, and output label. Setting the minimum number of sensors option to two prevents all the false detections in the table from becoming false detections on a system level.
The false detections caused by insects flying just in front of the sensors are very short-lived. The ones caused by clouds can last longer, sometimes several seconds.
As described above, the individual weaknesses observed for the primary sensors are the sensitivity to clouds of the IRcam and the autofocus of the Vcam. However, running the whole detection system has also shown that such individual shortcomings can be overcome using a multi-sensor solution.

\section{Discussion and Conclusions}

The increased use of drones and the raising concerns of safety and security issues following from this highlights the need for efficient and robust drone detection systems.
This work explores the possibilities to design and build a multi-sensor drone detection system utilizing state-of-the-art machine learning techniques and sensor fusion.
The system incorporates common video and audio sensors, and a thermal infrared camera.
To steer these cameras to specific directions of interest, the system also incorporates a fish-eye lens camera with a wider field-of-view that is used to detect moving objects.
An ADS-B receiver allows to keep track of cooperative aircrafts in the surrounding airspace.
All these sensors are mounted on a pan/tilt platform that sits on a standard tripod. %, in a way that they can be disassembled into %a few large parts for easy transport and deployment %(Figure~\ref{fig:system}).
%%
%All hardware and software elements are controlled with an %standard laptop that is also used to present results on the %screen (Figure~\ref{fig:system_GUI}).

%
Our results confirm that general machine learning techniques can be applied to input data from infrared sensors, making them well suited for the drone detection task. The infrared detector achieves a F1-score of 0.7601, showing similar performance as the visible video detector with a F1-score of 0.7849. The audio classifier achieves a F1-score of 0.9323.
Besides the analysis of the feasibility of an infrared sensor, this work expands the number of target classes utilized in the detectors compared to related papers. This work also includes a novel investigation of the detection performance as a function of sensor-to-target distance, with a distance bin division derived from the Detect, Recognize and Identify (DRI) requirements based on the Johnson criteria.
Due to the lack of a publicly available dataset, another main target has been to contribute with a multi-sensor dataset, which has been made publicly available. This dataset is especially suited for the comparison of infrared and visible video detectors due to the similarities in conditions and target types in the set.
To the best of our knowledge, this work is also the first to explore the benefits of including ADS-B data to better separate targets prone to be mistaken for drones.

%It has been observed that

One aspect not explored is the use of the Fcam together with the audio classifier as a means to output the position and label of a system detection.
Implementing a YOLOv2-detector on the Fcam could also be considered. However, a dataset to train it must either be collected separately or by skewing images from the visible video dataset, so that the typical distortion of a fish-eye camera is matched.
Neither is the performance of the audio classifiers performance as a function of sensor-to-target distance explored in the same way as the IR and visible sensors.
There is also a need for a better method to evaluate the whole drone detection system. To assess the performance of individual sensors is pretty straight forward, and in most cases, there are also numerous previous research to relate to, except for the IR-sensor. On a system level, however, existing research is very sparse, making any comparison difficult

As we have mentioned, most false detections are caused by either insects or clouds. Adding these classes to the dataset might be a way to overcome this. Using three different quadcopter drones makes the system effective against such drones. Extending the drone dataset by including also hexacopters, octocopters, fixed-wing and single rotor drones would be also useful before deploying the system to a real application.
It would also be of interest to use YOLO v3 instead, since it is more efficient in detecting small objects according to \cite{Unlu19}.
Further research could be to implement a distance estimation function to the target based on the output from the detectors.

The work done here is also applicable to other areas. One such that immediately springs to mind, is road traffic surveillance. Except for the ADS-B receiver, all other parts and scripts could be adopted and retrained to detect and track pedestrians or just a specific vehicle type, e.g. light ones such as motorcycles, or heavy ones like trucks.

\section*{Acknowledgment}

This work has been carried out by Fredrik Svanström in the context of his Master Thesis at Halmstad University (Master's Programme in Embedded and Intelligent Systems). The thesis is available at http://urn.kb.se/resolve?urn=urn:nbn:se:hh:diva-42141.  Author F. A.-F. thanks the Swedish Research Council and VINNOVA for funding his research.


%\section*{Acknowledgment}



\bibliographystyle{IEEEtran}

%\begin{thebibliography}{1}
%H.~Kopka and P.~W. Daly, \emph{A Guide to \LaTeX}, %rd~ed.\hskip 1em plus
% 0.5em minus 0.4em\relax Harlow, England: %Addison-Wesley, 1999.
%
%\end{thebibliography}

%\bibliography{bibliography}


\end{document}